\documentclass{article}

\usepackage{arxiv}

\usepackage[utf8]{inputenc} 
\usepackage[T1]{fontenc}    
\usepackage{hyperref}       
\usepackage{url}            
\usepackage{booktabs}       
\usepackage{amsfonts}       
\usepackage{nicefrac}       
\usepackage{microtype}      
\usepackage{graphicx}
\usepackage{natbib}
\usepackage{doi}

\usepackage{multirow}

\title{Amazon's 2023 Drought: Sentinel-1 Reveals Extreme Rio Negro River Contraction}

\author{ \href{https://orcid.org/0000-0002-9623-1182}{\includegraphics[scale=0.06]{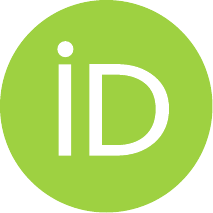}\hspace{1mm}Fabien H.~Wagner}\\
	Institute of the Environment and Sustainability,\\
	University of California, Los Angeles, CA 90095 USA;\\
	and	NASA-Jet Propulsion Laboratory\\
	California Institute of Technology, Pasadena, CA 91109, USA; \\ 
	and CTREES.org, Pasadena, United States;\\
	\texttt{wagner.h.fabien@gmail;fhwagner@ucla.edu} \\
 		\And
 	\href{https://orcid.org/0000-0002-8600-1949}{\includegraphics[scale=0.06]{orcid.pdf}\hspace{1mm} Samuel Favrichon} \\
	NASA-Jet Propulsion Laboratory\\
	California Institute of Technology, Pasadena, CA 91109, USA ; \\ 
	\texttt{samuel.favrichon@jpl.nasa.gov} \\
  \And
 	\href{https://orcid.org/0000-0002-7151-8697}{\includegraphics[scale=0.06]{orcid.pdf}\hspace{1mm}Ricardo Dalagnol} \\
	Institute of the Environment and Sustainability,\\
	University of California, Los Angeles, CA 90095 USA;\\
	and	NASA-Jet Propulsion Laboratory\\
	California Institute of Technology, Pasadena, CA 91109, USA ; \\ 
	and CTREES.org, Pasadena, United States;\\ 
 	\texttt{ricds@hotmail.com} \\
 		\And
 	\href{https://orcid.org/0000-0002-1817-360X}{\includegraphics[scale=0.06]{orcid.pdf}\hspace{1mm}Mayumi CM~Hirye} \\
 	Quapá Lab\\
 	Faculty of Architecture and Urbanism\\
 	University of São Paulo---USP\\
 	São Paulo, SP, Brazil\\
 	\texttt{ma.hirye@alumni.usp.br} \\
 	 	 	\And
 	  	\href{https://orcid.org/0000-0001-7711-8523}{\includegraphics[scale=0.06]{orcid.pdf}\hspace{1mm}Adugna Mullissa} \\
 	Institute of the Environment and Sustainability,\\
	University of California, Los Angeles, CA 90095 USA;\\
	and CTREES.org, Pasadena, United States;\\
 	\texttt{amullissa@ctrees.org} \\
		\And
 	\href{https://orcid.org/0000-0001-8524-4917}{\includegraphics[scale=0.06]{orcid.pdf}\hspace{1mm}Sassan Saatchi} \\
	Institute of the Environment and Sustainability,\\
	University of California, Los Angeles, CA 90095 USA;\\
	and	NASA-Jet Propulsion Laboratory\\
	California Institute of Technology, Pasadena, CA 91109, USA \\ 
 	\texttt{sasan.s.saatchi@jpl.nasa.gov}; \\ 
	and CTREES.org, Pasadena, United States
}

\renewcommand{\shorttitle}{\textit{arXiv} Template}

\begin{document}
\maketitle
\begin{abstract}
The Amazon, the world's largest rainforest, faces a severe historic drought. The Rio Negro River, one of the major Amazon River tributaries, reaches its lowest level in a century in October 2023. Here, we used a U-net deep learning model to map water surfaces in the Rio Negro River basin every 12 days in 2022 and 2023 using 10 m spatial resolution Sentinel-1 satellite radar images. The accuracy of the water surface model was high with an F1-score of 0.93. The 12 days mosaic time series of water surface was generated from the Sentinel-1 prediction. The water surface mask demonstrated relatively consistent agreement with the Global Surface Water (GSW) product from Joint Research Centre (F1-score: 0.708) and with the Brazilian Mapbiomas Water initiative (F1-score: 0.686). The main errors of the map were omission errors in flooded woodland, in flooded shrub and because of clouds. Rio Negro water surfaces reached their lowest level around the 25th of November 2023 and were reduced to 68.1\% (9,559.9 km$^2$) of the maximum water surfaces observed in the period 2022-2023 (14,036.3 km$^2$). Synthetic Aperture Radar (SAR) data, in conjunction with deep learning techniques, can significantly improve near real-time mapping of water surface in tropical regions.
\end{abstract}


\section{Introduction}


The Rio Negro River is among the largest tributaries of the left bank of the Amazon River, extending from its sources in Colombia to Venezuela and Brazil, making the Amazon the largest river basin in the world. The Rio Negro River has a length of 2230 km, a catchment area of 696,000 km$^2$, and an average flow of 28,400 m$^3$ per second, representing 14\% of the annual average flow of the Amazon basin \citep{Alho2015, Filizola2009}. It participates in the complex hydrological processes of the Amazon River that play a key role in the water, energy, and carbon cycles and interact with the global climate system \citep{FassoniAndrade2021}. On the 26th of October 2023, the Rio Negro River water level measured at the Port of Manaus (Brazil) reached its lowest point since 1903, 12.7 m. The water level measurement at the port is currently the most reliable data for the Rio Negro River, and to date, no near-real-time data of the river height or extent derived from medium spatial resolution satellite observation (5-30 m) is easily available over the Rio Negro basin.

Generating maps of near-real-time water surfaces location and extents for the Amazon region remains challenging. The primary obstacle is the persistent cloud cover. Despite the availability of numerous medium spatial resolution multispectral satellites like Landsat, CBERS, Sentinel-2, or PlanetScope, which acquire data at frequencies of $\leq$ one month, the extensive cloud cover impedes the creation of near-real-time observations of the state of the river basin. As a result, most water surface products are generated based on a yearly or multi-annual frequency \citep{Pekel2016, Hess2015, MapBiomas2024, Souza2019, Adeli2020}.

Only Synthetic Aperture Radar (SAR), due to its cloud-penetrating capability, appears as the data of choice for near real-time monitoring of surface water in tropical regions with persistent clouds. Moreover, water surfaces of the Amazon River in Synthetic Aperture Radar (SAR) images are easily distinguishable from other land covers due to their characteristic lower backscattering values compared to most terrestrial surfaces. These lower backscattering values are likely attributed to the specular reflection and smoothness of the water \citep{Hess1995, Alsdorf2000}.

There is a significant body of literature on the study of water levels, extent, and wetland vegetation in the Amazon region using radar, employing various instruments (SIR-C, RADARSAT-2, ALOS PALSAR, and ALOS2 ScanSAR, PolSAR, ENVISAT, JERS-1), diverse bands such as X-, C-, and L-band, and different polarizations \citep{Hess1995, Alsdorf2000, Silva2008, Sartori2011, DaSilva2012, Arnesen2013, Hess2015, AlmeidaFurtado2016, Cao2018, Canisius2019}, as reviewed in \citep{Adeli2020} and \citep{FassoniAndrade2021}. However, none of these studies provided near real-time and high-resolution water surfaces because the instruments or missions were not designed for this purpose. The Sentinel-1 C-band SAR satellite of the the European Space Agency (ESA), with its instrument operating at a center frequency of 5.405 GHz, enables image acquisition even in cloudy conditions. Furthermore, with a spatial resolution of 10 m, a temporal resolution of 12 days, and an open data policy, it currently appears as the optimal satellite program for measuring near-real-time variations in water surfaces at a fine scale and has not been fully exploited in the Amazon \citep{FassoniAndrade2021}.

To map water surfaces with Sentinel-1, several traditional methods are available, such as spectral indices \citep{Gao1996}, machine learning \citep{Huang2018a,Huang2021,Tang2022}, dynamic thresholding \citep{Bioresita2018, Xing2018, Uddin2019, Tiwari2020,Markert2020} and neural networks \citep{PhamDuc2017}. However, recently, the remote sensing scientific community is adapting to modern deep learning approaches \citep{Mayer2021}. For instance, the U-Net convolutional neural network can accurately map surface water from Sentinel-1 in tropical regions, achieving a high validation accuracy with an F1-Score exceeding 0.92 \citep{Mayer2021, Ronneberger2015}. Another study, which provides a substantial dataset of flooding samples for training models, demonstrates that a fully convolutional (FCNN) model outperforms thresholding algorithms in identifying flooded areas \citep{Bonafilia2020}. Further research indicates that the U-Net algorithm can learn features as meaningful as spectral indices for flood segmentation \citep{Konapala2021}. In an attempt to map floods with CNN-based methods and Sentinel-1 imagery with minimal pre-processing, it was shown that CNNs, including the U-Net algorithm, can reduce the time required to develop a flood map by 80\%, while maintaining strong performance across various locations and environmental conditions \citep{Nemni2020}. Lastly, for near-real-time flood mapping using Sentinel-1, the U-Net model demonstrated a notable improvement over thresholding techniques and enabled fast processing (1 min per image) with a very low omission error \citep{Katiyar2021}. Therefore, we used the U-Net convolutional neural network \citep{Ronneberger2015} to map water surfaces in Sentinel-1 images, as this algorithm is currently one of the most successful for mapping water in Sentinel-1 images and is an easy-to-use algorithm for binary segmentation tasks \citep{Konapala2021}.

Several products covering the Rio Negro River have been developed in the last decade. One of the most renowned and accurate water surface products is the Global Surface Water (GSW) at 30 m resolution from the Joint Research Centre (JRC) \citep{Pekel2016}. This dataset used the entire multi-temporal orthorectified Landsat 5, 7, and 8 archive spanning the past 32 years (1984-2016) and employed machine learning techniques to quantify global surface water and its changes at 30 m spatial resolution \citep{Pekel2016}. Specifically, using Google Earth Engine \citep{Gorelick2017}, they mapped water surface in three million Landsat satellite images, identifying when water was present, how occurrence changed, and the nature of changes in terms of seasonality and persistence. The main limitation of this product is that it is static and not available for recent years, thus cannot be used to monitor drought episodes in near-real time. Another remarkable static dataset for the region is the LBA-ECO LC-07 dataset of wetland extent for the Lowland Amazon Basin \citep{Hess2015}. This dataset was derived from mosaics of Japanese Earth Resources Satellite (JERS-1) Synthetic Aperture Radar (SAR) imagery for the period October–November 1995 and May–July 1996. It provides a map of wetland extent, vegetation type, and dual-season flooding state of the entire lowland Amazon basin at 3 arc-seconds of spatial resolution ($\sim$90 m). Finally, the most up-to-date and comparable to near-real-time dataset is the annual water surface from the MapBiomas initiative \citep{souza2020r,MapBiomas2024,Souza2019}, particularly from the year 2022. The water surface is mapped by the MapBiomas initiative using machine learning techniques and the Landsat Data Archive covering the period from 1985 to 2022 (scenes with cloud cover $<$ 70\%), available in the Google Earth Engine (Landsat 5 TM, Landsat 7 ETM+, and Landsat 8 OLI). This dataset is produced with $>$ 84\% average user accuracy \citep{souza2020r,Souza2019}. A limitation of the product is that the map is produced yearly, so it cannot be used for near-real-time mapping, for example, for the October - November 2023 drought. These three water surface datasets are the most accurate and used for our study region, and that is why we compared them with our results.

This work presents (i) the mapping of the water surface in the Rio Negro River basin at a 10 m spatial resolution and a 12 days temporal resolution using Sentinel-1 images and deep learning, (ii) the comparison of our water surface mask with the Joint Research Centre (JRC) Global Surface Water (GSW) at 30 m resolution, with the MapBiomas Water initiative water surface in 2022 at 30 m resolution, with the LBA-ECO LC-07 dataset of wetland extent and vegetation types and finally with the water level at the Port of Manaus, and (iii) description of the Rio Negro River contraction during the 2023 drought event. 

The water surface occurrence and recurrence on the 2022-2023 period are available at \url{https://doi.org/10.5281/zenodo.10552959}.

\section{Materials and Methods}

\subsection{Study site}

  \begin{figure}[ht]
 \centering
 \includegraphics[width=1\linewidth]{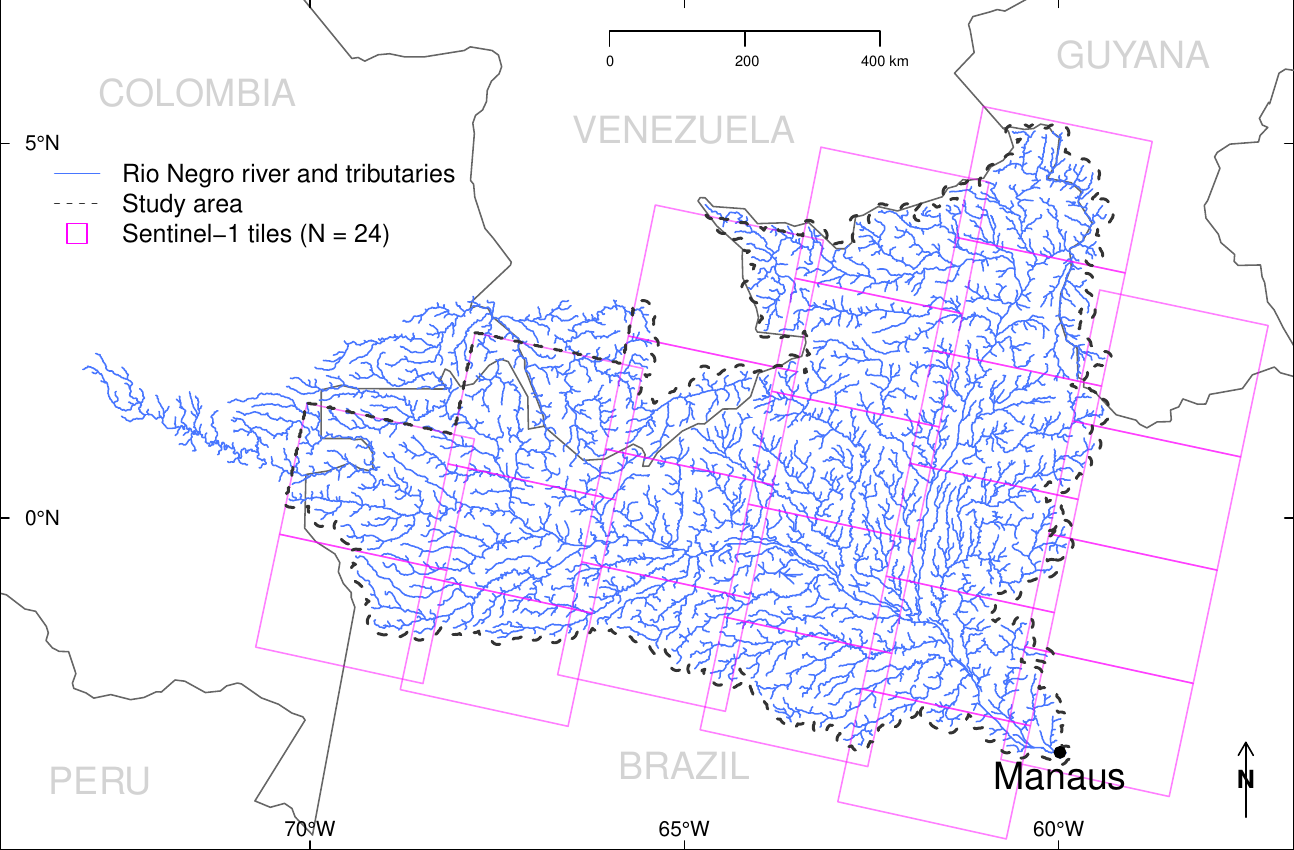}
  \caption{Geographical location in blue of the entire Rio Negro River and its tributaries before it forms the Amazon with the Solimões river at Manaus, adapted from \citep{FAO2022}. The study area is represented with a dashed black line, and the approximate extents of the 24 Sentinel-1 images taken for each orbit are in magenta.} 
  \label{Fig1}
  \end{figure}

\subsection{Sentinel-1 satellite images of the Rio Negro Basin}\label{satimg}

We used Sentinel-1 images from the global Sentinel-1 GRD archive at $\sim$ 10 m spatial resolution, covering the Rio Negro Basin in 2022 and 2023 in the Countries of Brazil, Venezuela, Guyana and Columbia (Fig \ref{Fig1}). These images were acquired in Interferometric Wide Swath (IW) mode with the VV and VH polarizations (dual-band cross-polarization, vertical transmit/vertical receive and vertical transmit/horizontal receive). Since the dataset was not contiguous over the basin, we retained only the largest contiguous part as our study area ($\sim$ 695,912 km$^2$, representing > 95\% of the basin area), Fig \ref{Fig1}. Sentinel-1 was accessed on 2023-12-24 and 2024-01-03 from \url{https://registry.opendata.aws/sentinel-1/}, and all images from the dates of 2022 and 2023 over the study area were downloaded (1536 images). 

To obtain calibrated and georeferenced images suitable for our water segmentation model, the VV and VH images in Geotiff were pre-processed using the SNAP-Python (snappy) API (SNAP, ESA Sentinel Application Platform v9.0.0, \url{http://step.esa.int}) of the Sentinel-1 Toolbox. The following steps were applied: (i) orbit correction to properly orthorectify the image using orbit metadata; (ii) GRD border noise removal to eliminate low-intensity noise and invalid data on scene borders, crucial for working with the entire scene; (iii) Thermal noise removal to minimize discontinuities between sub-swaths in scenes acquired using multi-swath acquisition modes; (iv) Calibration of the data to sigma-naught values, that is, computation of backscatter intensity with the sensor calibration parameters from the metadata; (v) Terrain correction (orthorectification) using the Global Earth Topography And Sea Surface Elevation at 30 arc-second resolution (GETASSE30) as the digital elevation model. The terrain-corrected values were converted to decibels via log scaling (10*log10(x)), and the VV and VH image values ranging between -49 to 1 were scaled to 0-255 for storage as an 8-bit integer 2-band raster, following the formula: (maximum(-49,minimum(1,x))+50)$\times$5.

\subsection{Sentinel-1 mountain shade mask from SRTM}
\label{mountaincor}

Topographic features create shadows in the SAR imagery that exhibit similar backscatter values to the water surfaces and may be misclassified by the model. Consequently, to exclude shaded areas from our results, we generated a shade mask  using the Shuttle Radar Topography Mission (SRTM) data at 30 m \citep{farr2007shuttle} to identify mountains and shaded slopes. The SRTM data for the study area were downloaded on 2023-12-29 from \url{https://dwtkns.com/srtm30m/}. The tiles were mosaicked into a single image, aggregated by a factor of 3 (90 m spatial resolution) using the minimum value using GDAL \citep{GDAL2023}. Then, slope computation was performed by considering the eight neighboring pixels with the R package \texttt{terra} \citep{Hijmans2023}. A value of 1 was assigned to slopes over 20 degrees, and 0 otherwise. The mask was then converted to polygons \citep{Pebesma2023,Pebesma2018}. Holes in the polygons were removed \citep{Dorman2023}, and the convex hull envelop was computed to ensure coverage of all mountains. Minor polygon errors (shaded areas over rivers) were manually edited. Finally, these polygons were rasterized into images matching the size and resolution of the images receiving the results of water segmentation.

\subsection{High resolution water masks and water levels}

To compare and analyze our water segmentation results, we used four datasets independent from Sentinel-1 observations over the region. The first product is the Joint Research Centre (JRC) Global Surface Water (GSW) at 30 m resolution based on Landsat data \citep{Pekel2016}. From the GSW product, we extracted data on water occurrence (percent of observation of the pixel as water in the time series, where 100 indicates constant water) and recurrence (the frequency of water reappearing from year to year across the time-series) over the Rio Negro basin. The datasets were accessed on 2023-12-21 from \url{https://global-surface-water.appspot.com/download}. The second product is the annual water surface of 2022 over our study area from the Mapbiomas Water initiative at 30 m resolution based on Landsat data \citep{souza2020r}. MapBiomas Project - Collection [8] of the Annual Land Use Land Cover Maps of Brazil, accessed on 2024-01-13 through the link: \url{https://brasil.mapbiomas.org/colecoes-mapbiomas/}. The water surface for the year 2022 (annual\_water\_coverage\_2022) was selected from the asset mapbiomas\_water\_collection2\_annual\_water\_coverage\_v1. The third product is the LBA-ECO LC-07 Wetland dataset \citep{Hess2015} accessed from \url{https://daac.ornl.gov/cgi-bin/dsviewer.pl?ds_id=1284} on  2024-12-01. The dataset consists of maps of wetland extent, vegetation type and dual-season flooding state of the entire lowland Amazon basin at 3 arc-seconds of spatial resolution ($\sim$90 m). These map were derived from mosaics of Japanese Earth Resources Satellite (JERS-1) Synthetic Aperture Radar (SAR) imagery for the period October-November 1995 and May-July 1996. Full classes and values description are available here: \url{https://daac.ornl.gov/LBA/guides/LC07_Amazon_Wetlands.html}. The fourth product is the daily Rio Negro water level measured at the port of Manaus (Brazil), accessed on Jan 6 from \url{https://www.portodemanaus.com.br/?pagina=nivel-do-rio-negro-hoje}.


\subsubsection{Model Architecture}

\begin{figure}[ht!]
\centering
\includegraphics[width=0.90\linewidth]{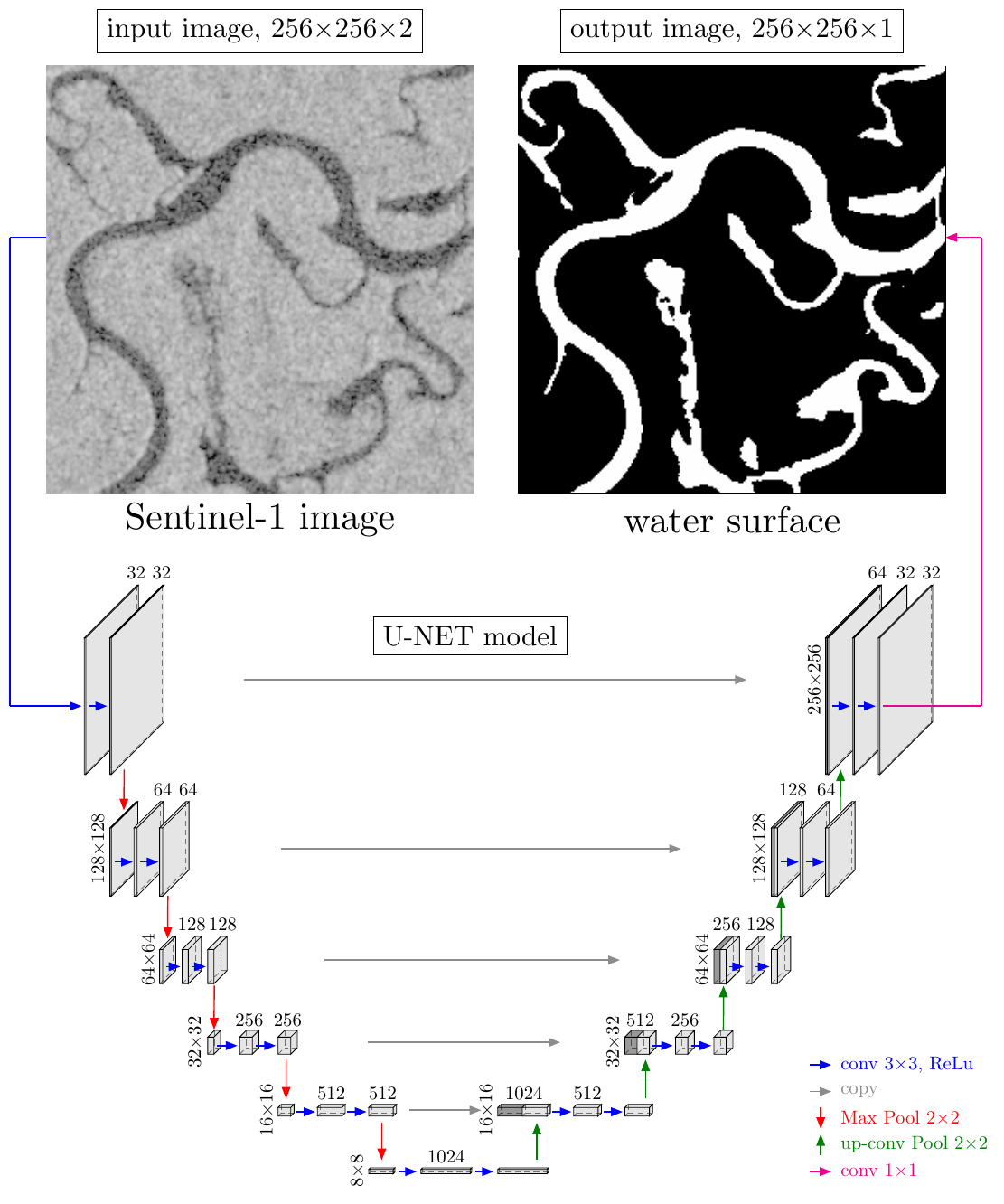}
\caption{U-Net model architecture used for water surface estimation from Sentinel-1 images, adapted from \citet{Ronneberger2015}. The number of channels is indicated above the cuboids, and the vertical numbers indicate the row and column size in pixels. The operations (convolutions, skip connections, max pooling and upsampling) performed in each layer and their sizes are indicated by the colored arrows.}
 \label{FigUnet} 
 \end{figure}

The water surface segmentation was performed using a classical U-net model \citep{Ronneberger2015}, Fig. \ref{FigUnet}. Specifically, the U-net model returns the probability of water surface presence in each pixel of a given input image. The model takes 2-band (VH and VV) Sentinel-1 images with a size of 256 $\times$ 256 pixels as inputs. The output is a one-band mask with dimensions 256 $\times$ 256 pixels, containing 1 (indicating water surface with a pixel probability $>=$0.5) or 0 (representing non-water pixel with a probability $<$0.5). The model was implemented in the R language \citep{CoreTeam2016} using the RStudio interface to Keras and TensorFlow 2.10 \citep{chollet2015keras,AllaireChollet,allaireTang,AbadiAgarwalBarhamEtAl2015}.

\subsubsection{Network training}

\subsection{Training}

To generate training samples for the U-net model's water surface, we used a water mask at 4.78 m resolution derived from a previous model using Planet NICFI \citep{Wagner2023MT,planet2021}. From this dataset covering Amazon water bodies from 2015 to 2023 and tiled with the Planet NICFI tiling system, we selected water masks from 2021 to 2022 that were fully covered by a Sentinel-1 image (acquired $\leq$ 25 days apart from the NICFI mosaic date) and had less than 25\% cloud cover. Then water masks were randomly selected for each monthly Planet mosaic date from 2021 and 2022, to constitute a data set of 550 non-overlapping images. Additionally, 29 Planet NICFI-derived water masks containing sandbanks in 2021 and 2022 were added to this dataset. This strategy was implemented to have images for all seasons, use only non-overlapping scenes to prevent overfitting, and include sandbanks observed solely in the dry season, Fig. \ref{Fig1}.

The Sentinel-1 images corresponding to the water mask tiles were pre-processed as in section \ref{satimg} and clipped to the water mask extents. Subsequently, the water mask at 4.78 m resolution was projected onto a raster with the same resolution and extent as the preprocessed and clipped Sentinel-1 image. Both the clipped Sentinel-1 image and the corresponding water mask, having the same size and resolution, were then cropped to a size of 1792 $\times$ 1792 pixels and retiled to the size of 256 $\times$ 256 images for use in model training. The complete sample comprises 59,387 and 6,057 Sentinel-1 image patches of 256 $\times$ 256 pixels and their associated water mask, for training and validation, respectively.



Each image patch underwent a data augmentation process, involving random vertical and horizontal flips. No additional data augmentation was necessary due to the natural data augmentation provided by the acquisition of sampling images at different dates \citep{Wagner2021}. Following data augmentation, the images were fed into the U-net model.

During network training, we used a standard stochastic gradient descent optimization. The loss function was designed as a~sum of two terms: binary cross-entropy and Dice coefficient-related loss of the predicted masks \citep{Dice1945,AllaireChollet,chollet2015keras} and finally the optimizer Adam \citep{Kingma2014} with a learning rate of 0.0001 was used. We used the accuracy (i.e. the frequency with which the prediction matches the observed value) as the metrics to assess the model performance.

The network was trained for 2000 epochs with a batch size of 32 images and the model with the best accuracy was kept for prediction. The training of the models took approximately 4.5 hours per 100 epochs using a NVIDIA A10G Tensor Core (GPU) with a 24 GB memory. 


 \subsection{Prediction}

For the prediction of the water mask, the 1536 Sentinel images (64 dates spanning 2021-12-02 to 2023-12-31 for each of the 24 Sentinel-1 tiles, Fig. \ref{Fig1}) were resized by adding columns and rows to have an aspect ratio of 4096 and a 128-pixel border. This adjustment was made to meet the input size requirements for prediction. The standardized-size Sentinel-1 tiles were then subdivided into sub-images of 4,224 $\times$ 4,224 pixels with a 128-pixel overlap between them. The total number of sub-images to predict was 69,577. Predictions were made on these 4,224 $\times$ 4,224 pixel images, and the 128-pixel border on each sub-image prediction was removed to mitigate border artifacts \citep{Ronneberger2015}.  To create the 11-day water mask mosaics, the resulting 4096 $\times$ 4096 images were projected onto a regular grid with a tiling system of 4096 $\times$ 4096 pixel images, maintaining the same resolution as the central scene of Sentinel-1 over the study area during each 11-day period. In cases of overlap between predicted images, the maximum value was retained, and in the absence of data, the previous mosaic values were used to complete the time series. The final dataset, covering the period 2022-2023, consisted of 62 mosaics with a temporal resolution of 12 days. The computation time for predicting Rio Negro water masks using an RTX4090 GPU was 3.75 days.

\subsection{Filtering for artifacts in Sentinel-1 images}
\label{cldcor}

Although the Sentinel-1 C-band SAR can theoretically penetrate clouds, some images may still exhibit significant artifacts, potentially from convective clouds as previously observed in this region \citep{Doblas2020b}. This leads to an increase in the values of the VV and VH bands, making it challenging for the model to detect water. To address this artifact in the prediction, we selected all tiles in our prediction grid with more than 500,000 pixels of river (68 tiles) based on a water occurrence mask computed for the entire time series (62 dates). For each of these 68 tiles, we identified the 5 images with the largest anomalies in pixels classified as water in the water occurrence mask and having fewer than two dates without water. Subsequently, all selected prediction tiles underwent visual inspection, and in cases where significant omissions of water pixels were clearly identified as errors, corrections were made using the previous water mask mosaic date (111 images were corrected).

\subsubsection{Segmentation accuracy assessment}
The F1 score was computed for each sample $i$ (training and validation) as the harmonic average of the precision and recall, eqn \ref{eqn1}, where precision was the ratio of the number of segments classified correctly as $i$ and the number of all segments (true and false positive) and recall was the ratio of the number of segments classified correctly as $i$ and the total number of segments belonging to class $i$ (true positive and false negative). This score varies between 0 (lowest value) and 1 (best value). 
\begin{eqnarray}
\label{eqn1}
F1_i& = & 2 \times \frac{precision_i \times recall_i}{(precision_i + recall_i)}
\end{eqnarray}

\section{Results}

\subsection{Model accuracy}

Overall, the F1-score, precision and recall values were similar for the training and validation sample, Table \ref{F1score}.
The water surface model F1-scores of the predictions on the training and validation samples were both high with a value of 0.930, Table \ref{F1score}. The recall was lower than the precision indicating a slightly higher rate of false negative than of false positive.

\begin{table}[ht]  
\centering
\caption{F1-scores of the water surface segmentation in the training and validation samples.}

\begin{tabular}{r|r|r|r|r|r}
model & sample & nb images &precision&recall& F1-score\\
\hline
\multirow{ 2}{*}{water mask}& training & 59,387 &   0.935 & 0.926 & 0.930 \\
&validation & 6,057 & 0.934& 0.926& 0.930 \\
    \hline
 \end{tabular}
\label{F1score}
\end{table}

\clearpage
\begin{figure}[ht]
\centering
\includegraphics[width=0.90\linewidth]{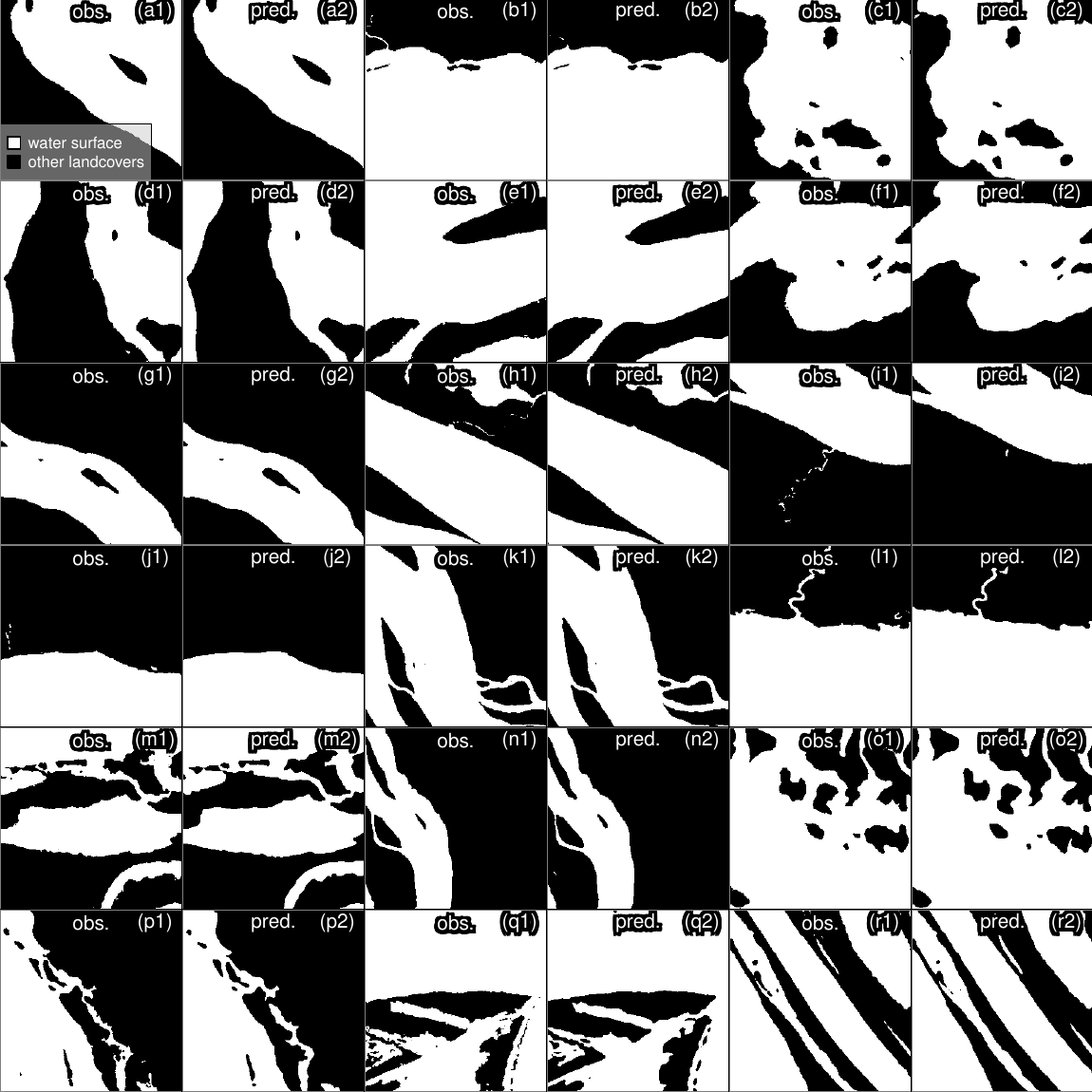}
\caption{Observation and prediction for 18 images of the validation dataset with a F1-score above 0.97.}
 \label{Figvalgoodf1} 
 \end{figure}

The model performs well on the validation dataset and can accurately capture most of the variations in the river border, Fig. \ref{Figvalgoodf1}a1 - r2. It can segment well very small islands, Fig. \ref{Figvalgoodf1}f2 and g2. However, the resolution of the mask may be slightly degraded in certain areas, as seen in Fig. \ref{Figvalgoodf1}o2, p2, and q2. It is important to note that the water surface masks for training and validation come from a model at 5 m spatial resolution (and were aggregated at 10 m), while Sentinel-1 spatial resolution is 10 m. Overall, the model performs very well for non-water pixels and has very few commission errors. However, for very small rivers, there are some omissions, even in the validation images with the highest F1-score, where small rivers with one or two pixels in width (10-20 m) are missed, Fig. \ref{Figvalgoodf1}b2, h2, i2, and j2.

\begin{figure}[ht!]
\centering
\includegraphics[width=0.90\linewidth]{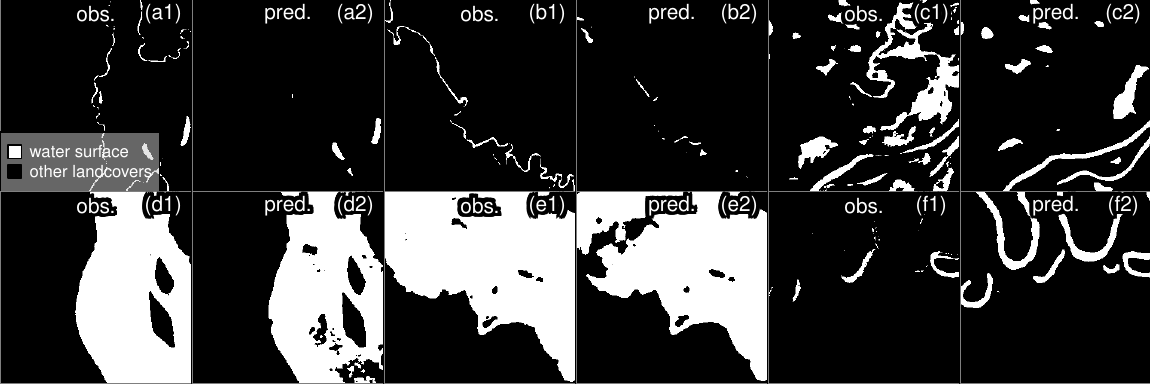}
\caption{Observations and predictions for six images of the validation dataset with an F1-score below 0.5 or cloud artifacts.}
 \label{Figvalbadf1} 
 \end{figure}

For the images in validation with low F1-scores, errors can arise from the omission of small rivers with widths less than 10 to 20 m, Fig. \ref{Figvalbadf1}a1 - b2. An inundation event in flooded shrubs was also missed by the model, Fig. \ref{Figvalbadf1}c1 - c2. Errors from clouds are visible in \ref{Figvalbadf1}d2 and e2. These cloud errors are recognizable as their borders are uncorrelated with the river pattern and usually present in the entire Sentinel-1 image. Finally, some errors are due to clouds in the Planet NICFI-derived water surface mask, Fig. \ref{Figvalbadf1}f2, while water surface is correctly mapped from the Sentinel-1 image.

Concerning the errors observed in the prediction, common omission errors include the impact of clouds in the Sentinel image, resulting in an increase in VH and VV bands, leading to the omission of water. We have filtered out most of these large errors in post-processing (see section \ref{cldcor}). The band of water directly in front of the city of Manaus exhibits some omission errors, but it is challenging to determine if it comes from the raw image, image correction, or proximity to very high VH and VV values in the city that could influence the model. This issue may be mitigated in the future near cities by incorporating training samples in those areas. In our study, this affects a negligible area in relation to the total water surface of the Rio Negro River. Regarding commission errors, all mountain-shaded areas are predicted as water, necessitating masking through alternative means, as they exhibit the exact same pattern as water (see section \ref{mountaincor}). Additionally, a less significant commission error in terms of area over our study area but consistently present is airport landing runways.

\subsection{LBA-ECO LC-07 Wetland dataset}

Regarding the vegetation types and seasonal flooding states (wetland LC-07 data) of the pixels classified as water surface by our model, Table \ref{tab:res}, 29.7\% fall into the open water class in the wetland LC-07 data. The next most frequent classes were "Non-flooded shrub/Flooded shrub" (12.7\%), Non-wetland within Amazon Basin (12.0\%), "Non-flooded forest/Flooded forest" (11.3\%), and "Flooded woodland/Flooded woodland" (8.1\%). The high percentage of pixels classified as "Non-wetland within Amazon Basin" (12.0\%) is likely due to the higher spatial resolution (10 m) of our dataset, which maps more smaller rivers than the LC-07 wetlands dataset (90 m spatial resolution).

\begin{table}[ht]
\small
\caption{Frequency and percentage of Wetland classes for the pixels classified as water by our model and for the pixels classified as non-water by our model but classified as water surface in the GWS dataset (false negative).}
\centering
\hspace*{-1cm}
\begin{tabular}{lllll}
  \hline
&   & Our model & False negative & False negative \\ 
\multicolumn{2}{c}{Vegetation type and dual-season flooding state from LBA-ECO LC-07 data}   &  & GWS & MapBiomas \\ 
Cover at Low Water Stage & Cover at High Water Stage & Freq. (\%) & Freq. (\%) & Freq. (\%) \\ 
  \hline
Non-wetland within Amazon Basin & Non-wetland within Amazon Basin & 3040627 (12) & 310982 (10.2) & 743987 (9.6) \\ 
  Open water & Open water & 7564727 (29.7) & 338274 (11.1) & 256448 (3.3) \\ 
  Open water & Aquatic macrophyte (flooded herbaceous) & 214580 (0.8) & 8275 (0.3) & 22046 (0.3) \\ 
  Non-flooded bare soil or herbaceous & Open water & 894024 (3.5) & 317612 (10.4) & 392493 (5.0) \\ 
  Non-flooded bare soil or herbaceous & Aquatic macrophyte (flooded herbaceous) & 590643 (2.3) & 199393 (6.5) & 343246 (4.4) \\ 
  Aquatic macrophyte (flooded herbaceous) & Aquatic macrophyte (flooded herbaceous) & 366235 (1.4) & 121902 (4) & 223821 (2.9) \\ 
  Non-flooded shrub & Open water & 48268 (0.2) & 2645 (0.1) & 1290 (0.0) \\ 
  Non-flooded shrub & Flooded shrub & 3239111 (12.7) & 499891 (16.4) & 1700976 (21.9) \\ 
  Flooded shrub & Open water & 496599 (2) & 206689 (6.8) & 318729 (4.1) \\ 
  Flooded shrub & Flooded shrub & 25446 (0.1) & 3748 (0.1) & 4115 (0.1) \\ 
  Non-flooded woodland & Flooded woodland & 501838 (2) & 30312 (1) & 17367 (0.2) \\ 
  Flooded woodland & Flooded woodland & 2059197 (8.1) & 619445 (20.3) & 2003905 (25.7) \\ 
  Non-flooded forest & Non-flooded forest & 1692959 (6.7) & 136088 (4.5) & 697255 (9.0) \\ 
  Non-flooded forest & Flooded forest & 2863932 (11.3) & 94924 (3.1) & 343367 (4.4) \\ 
  Flooded forest & Flooded forest & 1723735 (6.8) & 147859 (4.9) & 632938 (8.1) \\ 
  Elevation $>$= 500 m, in Basin & Elevation $>$= 500 m, in Basin & 105759 (0.4) & 10394 (0.3) & 81478 (1.0) \\ 
   \hline
\end{tabular}
\label{tab:res}
\end{table}

\subsection{Comparison with Mapbiomas Water Initiative data - year 2022}

We found 24,761,019 and 24,774,540 pixels, corresponding to $\sim$ 22,285 and $\sim$ 22,297 km$^2$ of water surface at 30 m resolution from our model in 2022 and from the Map Biomass water surface in 2022, respectively. The F1-score between water presence/absence in the Mapbiomas dataset and our water surface at the same spatial resolution (30 m) was 0.686. The precision was 0.686, indicating a high rate of false positives, which was expected as our dataset is produced at a higher resolution (10m), allowing us to capture more rivers. The recall was 0.686, similar to the precision value, indicating a significant rate of false negatives. Main differences can be analyzed with the vegetation and seasonal states of the pixels classified as false negative, Table \ref{tab:res}. Almost half of the false negatives fell in the classes annually flooded woodland (25.7\%) and seasonally flooded shrub (21.9\%). 


\subsection{Comparison with Global Water Surface (JRC) - period 1984–2015)}

We found 25,428,448 and 18,650,053 pixels ($\sim$ 22,886 and $\sim$16,785 km$^2$) of water surface at 30 m resolution from our model in the period 2022-2023 and the GWS dataset, respectively. The F1-score between water presence/absence in the GWS dataset and our water surface at the same spatial resolution (30 m) was 0.708. The precision was 0.614 in relation to the GWS dataset, indicating a high rate of false positives, as expected since our dataset is produced at a higher resolution (10 m) allowing us to capture more rivers. The recall was 0.836, indicating a relatively low rate of false negatives. As for the comparison with MapBiomas water surface, a product that is also based on Landsat data, the two prominent classes in the false negatives were annually flooded woodland (20.3\%) and seasonally flooded shrub (16.4\%).



\subsection{Regional results and 2023 drought at 10 m spatial resolution}

\begin{figure}[ht]
\centering
\includegraphics[width=0.90\linewidth]{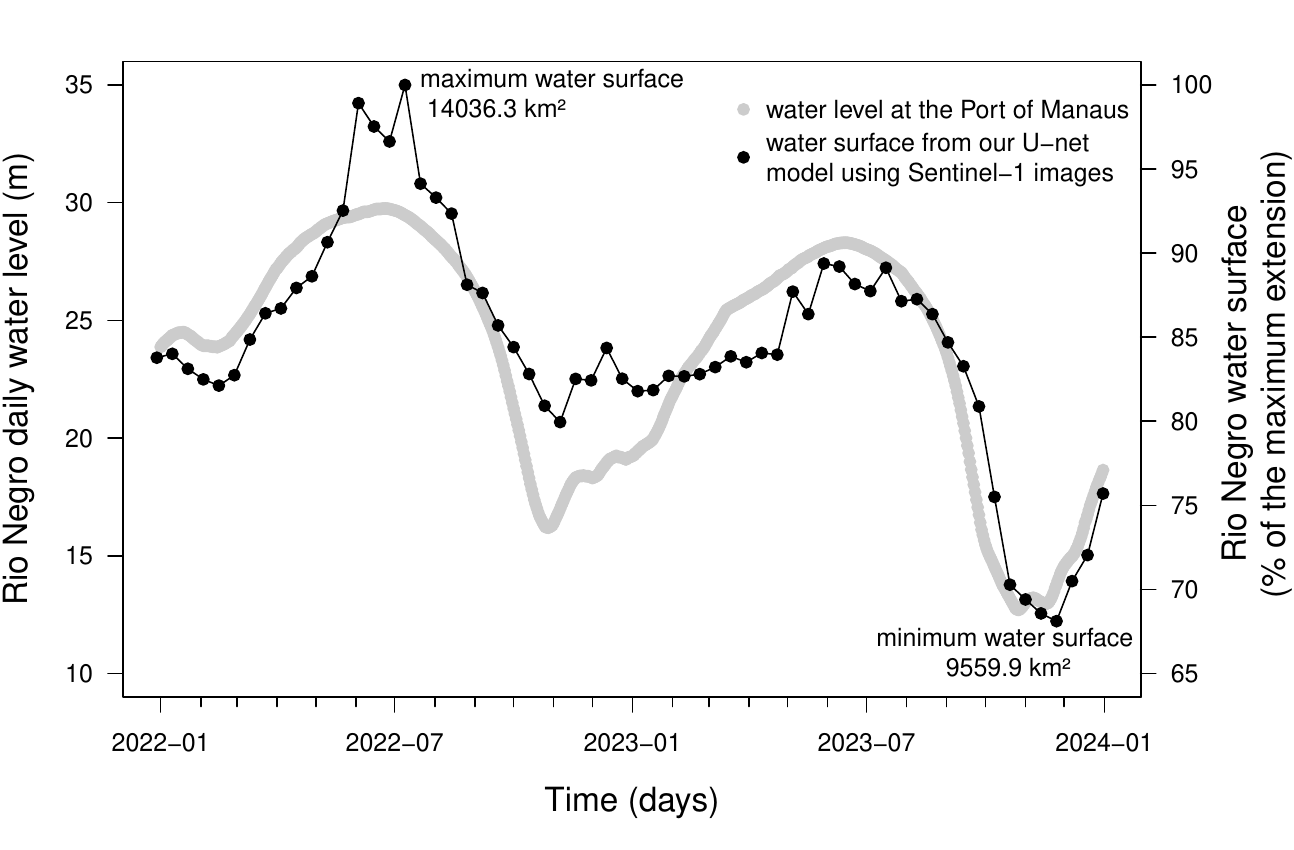}
\caption{Daily water level of the Rio Negro measured at the Port of Manaus and water surface estimated every 12 days with our Sentinel-1 based model for the Rio Negro study area, for the period 2022-2023.}
 \label{FigPort} 
 \end{figure}

The correlation between the Rio Negro River water levels measured at the Port of Manaus and the water area for the basin, measured from our model and Sentinel-1 images, was high, with a value of 0.887, as shown in Fig. \ref{FigPort}. The minimum and maximum values of both curves always occurred with a maximum delay of $\sim$ one month. For the highest water level around July, the water surface is more variable, likely due to large new water surfaces from inundation. For the lowest water levels, both curves exhibit a similar pattern of variation, suggesting that water surfaces are more associated to the river's water level during the dry period of the year.

The median water surface area for the 2022-2023 period was 11,795.2 km$^2$. The maximum water surface was observed on July 9, 2022, reaching an area of 14,036.3 km$^2$. The lowest water surface, 9,559.9 km$^2$, was recorded on November 25, 2023, in the water surface mosaic composed of Sentinel-1 images taken from the 14th to the 25th. During the 2023 drought, the water surface decreased to 68.1\% of the maximum observed and 81.0\% of the median observed during the period 2022-2023.

\begin{figure}[ht]
\centering
\includegraphics[width=1.00\linewidth]{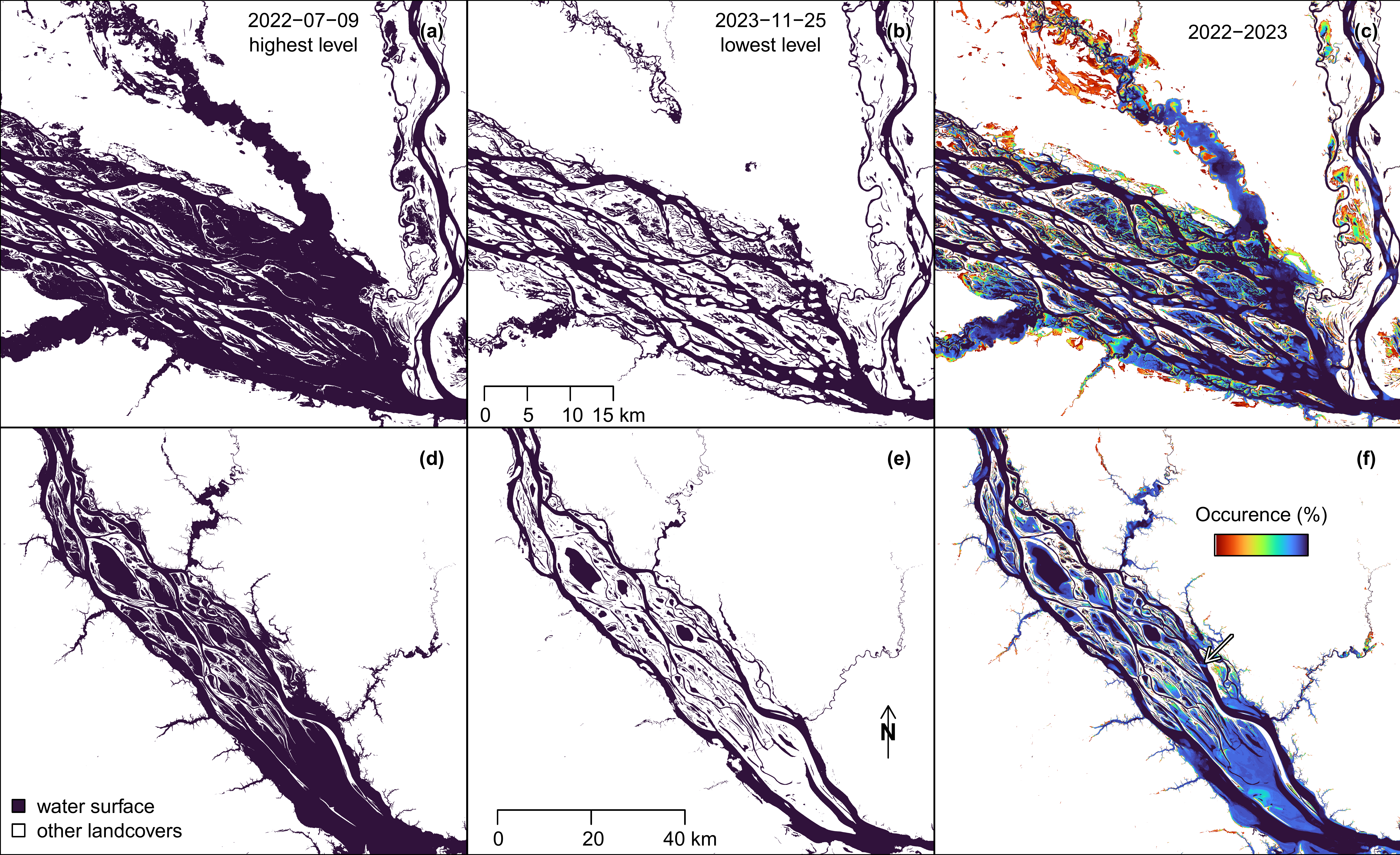}
\caption{Examples of our deep learning water segmentation results based on Sentinel-1 images for the Rio Negro water surface near the junction with the Branco River (a-c) and in near the Anavilhanas Archipel National Park (d-f), for the maximum observed (2022-07-09, first column), the minimum observed (2023 drought, 2023-11-25, second column) and water occurence per pixel in \% computed on the 2022-2023 period (third column). A white arrow in (f) indicates omission errors at the border between two images in the same orbit.} 
 \label{FigWetDry} 
 \end{figure}

Two of the major archipelagos in the Rio Negro River, near the junction with the Rio Branco river and the Rio Jufari river, Fig. \ref{FigWetDry}a-c, and near the Anavilhanas Archipelago National Park, Fig. \ref{FigWetDry}d-f, showed a remarkable pattern of changes between the highest and lowest water surfaces estimated by our model. During the highest measured water surface, some islands are still above water but the river dominates the landscape, with 1055.6 km$^2$ for the Rio Branco junction Fig. \ref{FigWetDry}a and 1711.6 km$^2$ for the Anavilhanas Archipelago, Fig. \ref{FigWetDry}d. During the lowest measured water surface on November 25, 2023, the water surfaces were reduced by half in both archipelagos, with 549.5 km$^2$ (52.1\% of the highest level) for the Rio Branco junction and 841.2 km$^2$ (49.1\% of the highest level). The part of the river that disappears in \ref{FigWetDry}b still has some water during the drought, as verified in Planet NICFI images at 5 m resolution for October and November. However, it remains undetected by the model for the date of the lowest point and the three previous dates, and the river becomes connected again the date after the lowest point. Also, the VV and VH bands look more similar to sand banks than to water. This could be explained by thin water paths that are undetected, very low water levels affecting radar measurements (shoaling), or artifacts of the GRD border noise removal to eliminate low-intensity noise and invalid data on scene borders in the image, as this river is on the overlap of two Sentinel-1 orbits.

For both the junction with the Rio Branco river and the Anavilhanas Archipelago, 40\% of the water surface is permanent (occurrence of 100\%), Fig. \ref{FigWetDry}c and f. However, the seasonality differs, as the value of the 20th percentile for water occurence at the river junction with the Rio Branco river is 54\%, while for Anavilhanas Archipelago, it is 85\%. This indicates less seasonality in water surface for the Anavilhanas Archipelago, and there is more possibility of inundation at the junction with the Rio Branco river as seen with reddish colors in \ref{FigWetDry}c. An artifact indicating omission errors at the border between two images in the same orbit is highlighted by a white arrow in Fig. \ref{FigWetDry}f.  There is no overlap between images captured in the same orbit and this artifact is likely attributed to a border effect inherent in the raw Sentinel image.

%

\section{Discussion}

\subsection{Mapping water surface of the Rio Negro River and perspectives}

Here, for the first time, we demonstrate that medium spatial resolution Sentinel-1 C-band SAR allows the accurate estimation of water surface extent and changes of the Rio Negro River in near-real-time, with a 12-day temporal resolution and 10 m spatial resolution. The water surface segmentation model achieved an F1-score above 93\% on the validation dataset, highlighting once again the high capacity of deep learning to support water mapping in tropical environments \citep{Mayer2021,Bonafilia2020,Konapala2021,Nemni2020}. The good performance of the segmentation could be explained by the unique water surface low backscattering value in comparison to other land covers \citep{Hess1995,Alsdorf2000}. Our model enables accurate near-real-time monitoring of the water surfaces and performed particularly well during drought. For the period 2022-2023, we estimated a median water surface area of 11,795.2 km$^2$, a maximum of 14,036.3 km$^2$, and a minimum of 9,559.9 km$^2$, observed during the 2023 drought, Fig. \ref{FigPort}. In this major drought, with water levels reaching the lowest observed since 1903, the water surface decreased to 68.1\% of the maximum observed and 81.0\% of the median observed during the period 2022-2023. When water levels are at their lowest during the year, water level and water surface exhibit a similar pattern of variation, suggesting that water surfaces are more associated with the river’s water level during the dry period of the year. Climate extreme events, such as major floods and droughts, have become more frequent in the Amazon region in the last decades \citep{Gloor2013,Marengo2016,Silva2023}, and our model could be used to monitor near-real-time changes in water extents in this region. Furthermore, in combination with recent data from the SWOT mission, which observes surface water storage change and fluxes at the global scale \citep{Biancamaria2016}, it could be used to model the real water balance of the Amazon River, in addition to its water surface extent. Finally, in a broader context, as flooding affects large regions in the world each year \citep{Lin2019,Uddin2019}, our model could be tested for near-real-time mapping of flood events.

\subsection{Limitations of the water surface model}

The model made very few commission errors after masking for mountain shades. Even though it does not represent a significant area, airport runways and large roads ($\sim$ four lanes) were predicted as water surfaces, likely due to low backscattering values in either polarization, resulting from specular reflection over smooth surface, as observed in SAR images \citep{Kumar2021}. Most mountain shades were predicted as water surfaces due to similar backscattering values. We could have used SRTM at 30 m spatial resolution for terrain correction, which is the conventional way. However, it was faster to process the terrain correction of the 1536 images using the Global Earth Topography And Sea Surface Elevation at 30 arc-second resolution (GETASSE30) digital elevation model. Furthermore, most rivers were located at low altitudes and far from mountain relief that could have caused shadows. Opting for an independently created and manually corrected shadow mask was faster, taking under half a day, than performing SRTM terrain correction at 30 meters in preprocessing and addressing shadows within the model. Finally, even if there are still a few shadows present on the map, they are not supposed to change in area during the year and impact the seasonality of the water surface, as acquisitions are the same throughout the year for an orbit.

The omission errors were more common than commission errors, and we identified two main sources: clouds and seasonally flooded shrubs and trees. The Sentinel-1 C-band SAR can penetrate most clouds, but some images (111) showed significant artifacts, potentially from convective clouds, as already observed in the Amazon region \citep{Doblas2020b}. This leads to an increase in the values of the VV and VH bands, making it challenging for the model to detect water. Unlike multi-spectral satellite images, clouds are not easily detected in the VH and VV bands, and only the large errors could be detected and corrected when they were visible in the prediction mask (section \ref{cldcor}). This is likely the major problem of detecting water surface in near-real-time. The second main source of omission errors was flooded vegetation, with $\geq$ 35\% of the omission errors located in the classes 'annually flooded woodland' and 'seasonally flooded' of the LBA-ECO LC-07 vegetation type and dual-season flooding state dataset, Table \ref{tab:res}. This might indicate that Sentinel-1 C-Band is sensitive to the remaining vegetation structure above water during the flooded period. These omission errors are mostly located in the Rio Negro interfluvial wetlands, a region found 200 km North/North-West of the junction of Rio Negro and Rio Branco rivers only flooded during the wet season \citep{Hess2015, FassoniAndrade2021}. While this could be a problem for inundation mapping, it has no effect on estimating the river contraction during a drought. Studies that classify flooded forests and floodplain lakes with emergent shrubs with radar usually use more SAR bands and polarisations, not only C-band and VV or VH polarization, to benefit from the radar double-bounce returns from water and vegetation surfaces \citep{Alsdorf2000,Hess1995,Wang1995}. In 2024, the NISAR mission will be launched, providing SAR data (L-band, 24 cm wavelength, and S-band, 10 cm wavelength) distributed at 3 m spatial resolution and 12-day temporal resolution over land \citep{Rosen2021}. This new SAR data will likely complement the Sentinel-1 data to improve the mapping of flooded forests and floodplain lakes with emergent shrubs. There were few omissions in the water front of Manaus which could have originated from the raw image, insufficient training data near the city, image correction, or proximity to very high VH or VV values in the city that could impact the model prediction. While it was not significant in our study, it could have an impact in regions with heavily urbanized river borders. Finally, a few omissions were found in a band of approximately 200 m on the border between images of the same orbit (10 pixels of 10 m on each side). Contrary to neighboring orbits, these images had no overlap between them. This might be a border effect from the model, the raw data, or GRD correction. If it comes from the model, it might be resolved by filling the missing part of the image with the value in the neighboring image of the same orbit. Finally, the backscatter coefficients, especially the VV polarization, can also be influenced by wind-induced surface roughness over open water \citep{PhamDuc2017} and/or low water levels causing shoaling \citep{Bian2020}; and this might explain why water can disappear from the prediction, Fig. \ref{FigWetDry}b. However, it could also be due to a too small river width to be detected or artifacts from GRD border noise removal as this river is on the overlap of two Sentinel-1 orbits, and more investigation is needed.

\subsection{Water surface segmentation model application on larger scale}
While this study maps the water extent of one of the major Amazon river tributaries, it still represents only around 20\% of the water extent of the Amazon basin \citep{Souza2019}, and only during 2 years. For the application of this method on a larger scale, temporally or spatially, there are three main challenges. The first is the dataset; Sentinel-1 data does not provide a ready-to-use product for the user, and each image has to be preprocessed by SNAP, which takes a significant amount of time ($>$ 30 seconds per image) and limits the application on a larger scale for a single user on a single machine. The second limitation is the absence of a standard Sentinel-1 tiling scheme, making it challenging for the user to easily find all images corresponding to their region and date of interest without having to search on external databases, which can be problematic for large datasets. The third limitation is that in Google Earth Engine \citep{Gorelick2017}, the only place where the Sentinel-1 images are preprocessed and ready to use, the use of deep learning, while already tested and working \citep{Mayer2021}, would likely have a prohibitive cost for large regional/country scales. To conclude, users would benefit from an archive of ready-to-use preprocessed 12 days Sentinel-1 mosaics with a standard tiling scheme..






\vspace{6pt}

\section*{Author contributions}Conceptualization, F.H.W., S.F., R.D., M.C.M.H., A.M. and S.S.; methodology, F.H.W., S.F., R.D., M.C.M.H., A.M. an S.S.; software, F.H.W. and S.F. and A.M.; validation, F.H.W. and S.F.; formal analysis, F.H.W. and S.F.; investigation, F.H.W., S.F, R.D., M.C.M.H., A.M. and S.S.; resources, S.S.; data curation, F.H.W., S.F. and M.C.M.H. ; writing---original draft preparation, F.H.W., S.F., R.D., M.C.M.H., A.M. and S.S.; writing---review and editing, F.H.W., S.F., R.D., M.C.M.H., A.M. and S.S.; visualization, F.H.W.; supervision, S.S.; project administration, S.S.; funding acquisition, S.S. All authors have read and agreed to the published version of the manuscript.

\section*{Funding}
This research received no external funding. 

\section*{Data availability} 
The water surface occurrence and recurrence on the 2022-2023 period are available at \url{https://doi.org/10.5281/zenodo.10552959}. Sentinel-1 data are available at  \url{https://registry.opendata.aws/sentinel-1/}.



\section*{acknowledgments}
The authors are grateful to the Grantham and High Tide Foundations for their generous gift to UCLA and grants to CTrees for bringing
new science and technology to solve environmental problems. This work was partially
conducted at the Jet Propulsion Laboratory, California Institute of Technology under a contract (80NM0018F0590) the National Aeronautics and Space Administration (NASA).

\section*{Conflicts of interest}The authors declare no conflict of interest. The funders had no role in the design of the study; in the collection, analyses, or interpretation of data; in the writing of the manuscript; or in the decision to publish the~results.


\bibliographystyle{unsrtnat}
\bibliography{references}  






\end{document}